\pdfoutput=1

\documentclass[11pt]{article}

\usepackage[]{acl2023}

\usepackage{times}
\usepackage{latexsym}
\usepackage{ulem}

\usepackage[T1]{fontenc}

\usepackage[utf8]{inputenc}

\usepackage{microtype}
\usepackage{import}
\definecolor{mygreen}{rgb}{0.0, 0.5, 0.0}
\usepackage{todonotes} 

\usepackage{caption}

\title{Diversity Enhanced Table-to-Text Generation via Logic-Type Control}

\author{Yotam Perlitz, Liat Ein-Dor, Dafna Sheinwald, Noam Slonim, Michal Shmueli-Scheuer\\
  IBM Research AI \\
  yotam.perlitz@ibm.com, \{liate, dafna, noams, shmueli\}@il.ibm.com}

\begin{document}
\maketitle
\begin{abstract}

Generating natural language statements to convey logical inferences from tabular data (i.e., Logical NLG) is a process with one input and a variety of valid outputs. This characteristic underscores the need for a method to produce a diverse set of valid outputs, presenting different perspectives of the input data.
We propose a simple yet effective diversity-enhancing scheme that builds upon an inherent property of the statements, their logic-types, by using a type-controlled table-to-text generation model. 
We demonstrate, through extensive automatic and human evaluations over the two publicly available Logical NLG datasets, that our proposed method both facilitates the ability to effectively control the generated statement type, and produces results superior to the strongest baselines in terms of quality and factuality-diversity trade-off. 
\end{abstract}

\section{Introduction}
\label{intro}
Table-to-text (T2T) generation is the task of generating natural language statements to convey information appearing in tabular data.
This task is relevant in real-world scenarios including generation of weather forecasts~\cite{goldberg1994using}, sports results~\cite{wiseman2017rotowire}, and more. 

A statement generated from tabular data can be inferred based on different levels of information. These range from a value of a specific cell to the result of logical or numerical operations across multiple cells, such as the average value of a column, or a comparison between rows. 

\begin{figure}[ht]
\includegraphics[width=\columnwidth]{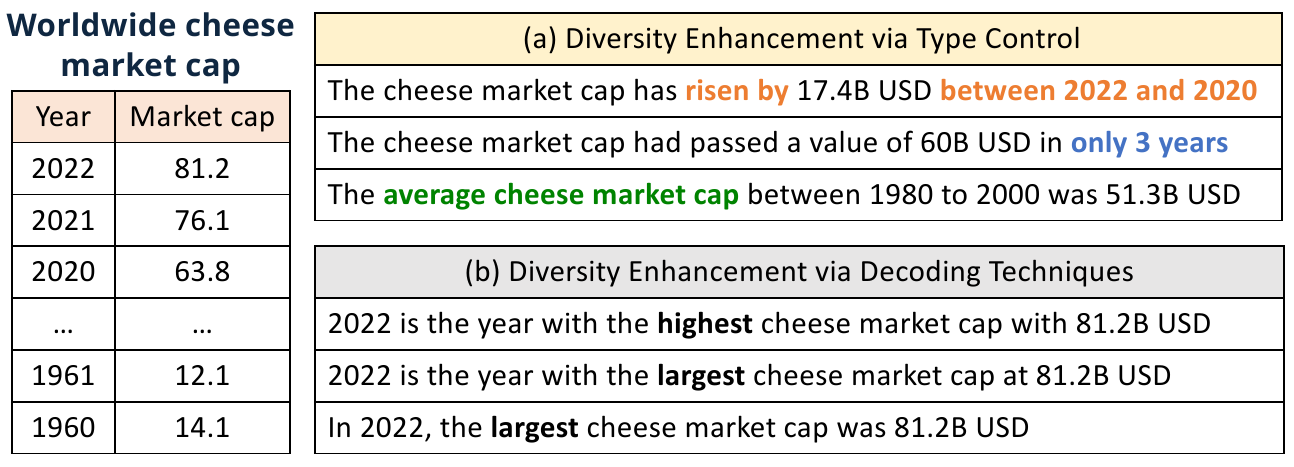}
\caption{
T2T generation of 3-statement sets for the table on the left; (a) LT controlled: each statement delivers a unique piece of information, yielded by the control employed: \textcolor{orange}{compare}, \textcolor{blue}{count}, and \textcolor{mygreen}{aggregation}; (b) decoding-based diversity: all are focused on one fact, hence demonstrating a weak diversity.}
\label{fig:motivation}
\end{figure}

In NLG in general, and in T2T generation in particular, a \textit{diverse set} of generated outputs given a single input is favorable, as it offers different perspectives on the data, provides the user with multiple options to choose from, and facilitates further improvement of output quality via post-generation re-ranking algorithms~\cite{gimpel2013systematic}. 

In this work, we propose a method for enriching the control and diversity of generated T2T outputs. To this end, we leverage a common semantic partition of T2T statements into \textit{numeric-logic types} (LTs)~\cite{chen2020logical} representing different perspectives of the data (see Figure~\ref{fig:motivation}(a)).

Namely, we utilize these LTs to realize a \textit{controlled} generation model, that allows guiding generated statements to a specific LT, out of the many different valid LTs corresponding to the input table. This controlled generation model enables our method, \textit{Diversity enhancement via LT Control} (\textsc{DevTC}) to produce a diverse set of statements representing multiple perspectives of the data, by conditioning upon several different LTs. 

\begin{figure*}[t]
\includegraphics[width=\textwidth]{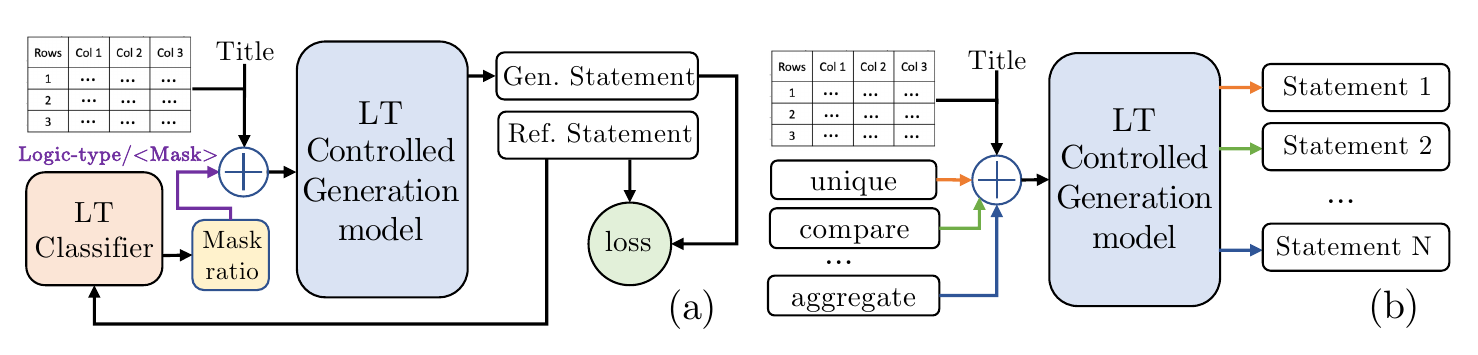}
\caption{Framework; (a) \textbf{Train}: the LT-conditional model is trained to 
generate 
a reference statement given the statement LT as it is predicted by our LT classifier; (b) \textbf{Inference}: \textsc{DevTC} is realized by inputting several different LTs along with a single table resulting in a diverse set of statements.}
\label{fig:model}
\end{figure*}


As previous T2T methods intrinsically can only produce a single output per input, they obtain output diversity through common decoding techniques that have been shown to suffer from a trade-off between diversity and quality measures such as fluency and  adequacy~\cite{ippolito2019comparison}. 
By this trade-off, high quality hinders diversity, as exemplified in Figure~\ref{fig:motivation}(b). 
In contrast, 
we show that \textsc{DevTC} readily generates a diverse set of high quality statements, allows LT-control, all while surpassing the baseline models in terms of generation quality.

Through extensive experimentation, we show that by employing this simple LT-control scheme, \textsc{DevTC} surpasses SOTA methods in the trade-off between diversity and quality, measured here in \textit{factuality} which is a paramount concern in T2T. 
We also show that \textsc{DevTC} generates statements adhering to the LT required by the user, and moreover, even in the absence of an input LT, outperforms the baselines on the two common benchmarks\footnote{Models and code will be made public upon acceptance.}.

\section{Related Work}
\label{related}

The task of \textsc{Logical NLG}, introduced by~\newcite{chen2020logical}, involves the automatic application of complex numeric-logic operations on data tables along with the  natural language expressions of them as statements. 
The task was accompanied by a dataset, \textsc{LogicNLG}, that contains a set of (table, statement) pairs. In addition to the \textsc{LogicNLG} dataset, \newcite{chen2020logical} presented two methods based on \textsc{GPT2}~\cite{radford2019language}. 
Both methods receive the same input $\mathbf{T}$: a table in conjunction with a title, denoted as a natural language sequence, but differ in their generation scheme. \textsc{GPT-TabGen} learns to generate a statement $Y$ directly: $p_{\theta}(Y|\mathbf{T})$; while \textsc{GPT-C2F} generates a statement-template, $\tilde{Y}$, and conditions on it to create the final statement, effectively learning $p_{\theta}([\tilde{Y};Y]|\mathbf{T})$. 
In a subsequent work, \newcite{chen2021confounded} proposed \textsc{DCVED}, a scheme based on a conditional variational auto-encoder architecture. Their scheme can generate multiple statements for a single input, but these only undergo a re-ranking, and their diversity or quality aspects are not discussed.
\textsc{Logic2Text}~\cite{chen2020logic2text} is a small dataset similar to \textsc{LogicNLG}. In its associated task, a model receives an additional logical-form input, specifying its full logical description. \newcite{liu2021improving} aims to circumvent the problem of data scarcity of \textsc{Logic2Text} with an approach combining data-augmentation, data-weighting and semi-supervised learning using LT-controlled generation module. In contrast to their work, our trained model is robust to missing LTs, and, paired with a diversity enhancing scheme, is shown to improve both generation diversity and factuality. 
Recently, \citet{zhao2023loft} successfully applied the proposed method to different tasks and models, extending it with additional LTs and a post-filtering module.

\section{Method}
\label{model}
\subsection{Statement-LT Classifier}
\label{classifier}
To enable controlled generation learning, we augmented our training datasets with LT-control annotations. 
Specifically, we automatically annotated our training datasets with $7$ LTs, namely, $c$ = \{count, comparative, superlative, unique, ordinal, aggregation, majority\} by employing a BERT~\cite{devlin2018bert} based classifier $p_{\phi}(c|Y)$ that was fine-tuned on $8.5$K (statement, LT) pairs from the \textsc{Logic2Text} train set.

This classifier achieved $97\%$ macro F1 on the corresponding test set. To measure the classifier's ability to transfer, we ran it on $200$ randomly sampled statements from \textsc{LogicNLG} annotated by experts achieving $90\%$ macro F1.

\subsection{LT-controlled T2T Generation Model}
As depicted in Figure \ref{fig:model}(a), we train an LT-controlled generation model, learning $p_{\theta}(Y|\mathbf{T},c)$, obtaining LT annotations from the reference statement using the statement-LT classifier.
The LTs are concatenated to the table and title to produce the input. 
The model is then trained to minimize the autoregressive cross-entropy loss between the generated and reference tokens. 

During training, we apply a mask over the LT with probability $p_{mask}=0.5$. 
Inducing an equal chance to receive a masked token as the LT in training, this ratio allows the model to learn how to condition on the LT, while also enabling robustness for scenarios where LT is unavailable for the model to condition on. 
The effects of other $p_{mask}$ choices are discussed in Appendix~\ref{A :masked_ratio}.

\subsection{Diversity Enhancement via LT Control}
Figure \ref{fig:model}(b) presents our \textsc{DevTC} inference-time flow. As the figure depicts, we utilize the above $p_{\theta}(Y|\mathbf{T},c)$ model to generate multiple statements, each conditioned on a different LT sampled from a uniform LT distribution. 
By this process, \textsc{DevTC} is able to produce statements with various types providing different perspectives on the data. 
\section{Experiments}
\label{experiments}

\subsection{Datasets}
In our experiments, we use \textsc{LogicNLG} \cite{chen2020logical} and \textsc{Logic2Text} \cite{chen2020logic2text}. 
Each data-point in \textsc{LogicNLG} consists of a parent-table crawled from Wikipedia from which $5$ tables are derived, each containing a subset of the parent-table columns and an associated statement generated by crowd-workers. 
\textsc{Logic2Text} is similar but further provides statement logical-form (its full logical description) from which we extract the LT. In our experiments, we will use these LTs to train a statement-LT classifier (cf. Section~\ref{classifier}) but will \textbf{not} use these extra annotations in training or evaluating the generation model. 
To the best of our knowledge, these two datasets are the only publicly available table-to-text datasets that include statement generation capturing complex logical and numerical operations from tables, making them the only datasets relevant for our scenario.

\subsection{Metrics}
\label{sec:metrics}
Following previously proposed evaluation practices laid out by~\citet{chen2020logical}, 
we evaluate the quality of a generated text, with BLEU 
to measure consistency with the reference text; 
and the \textsc{SP-Acc}  (SP-A)
and \textsc{NLI-Acc} (NLI-A) metrics to estimate its factuality, using semantic parsing and a pretrained NLI model, respectively.
Specifically, we focus on NLI-A that was found to better agree with human preference for factuality evaluation~\cite{honovich2022true}. 
For measuring the diversity of the generated statements we use the three common n-gram based metrics Self-BLEU$n$~\cite{zhu2018texygen},  Ent-$n$~\cite{zhang2018entk} and Dist-$n$~\cite{li-2016-dist}. 

\subsection{Hyper-parameters \& Baseline Methods}
We use the same hyper-parameters as in~\newcite{chen2020logical}, apart from the learning rate (LR) for which we tried $6$ values between $1e$-$6$ to $5e$-$5$ and chose the best LR per according to our model selection scheme, that uses the dev set BLEU3 score. 
As for models, we compare \textbf{\textsc{DevTC}} based on GPT2-small/medium with \textbf{\textsc{GPT-C2F}} and \textbf{\textsc{GPT-TabGen}}, and \textbf{\textsc{DCVED}}. \textsc{DCVED} is compared against the medium models only since it uses two GPT2-small and two fully-connected networks, adding up to a larger parameter count than GPT2-medium. 
Further details can be found in App. \ref{A: Implementation details}.

\label{results}
\section{Results}

\begin{table}[t]
\centering
\resizebox{\columnwidth}{!}{ 

\begin{tabular}{lcccc}
& \multicolumn{2}{c}{\textsc{LogicNLG}} & & \\ \hline
Model & Size & BLEU 1/2/3 ($\uparrow$) & SP-A ($\uparrow$) & NLI-A ($\uparrow$)  \\ \hline
\textsc{GPT-C2F} & sm & 46.6 / 26.8 / 13.3 & 42.7 & 72.2  \\

\textsc{GPT-TabGen} & sm & 48.8 / 27.1 / 12.6 & 42.1 & 68.7  \\

\textsc{DevTC$_{mask}$} & sm & \textbf{50.0} / \textbf{28.6} / \textbf{14.4} & \textbf{43.0} & \textbf{73.4} \\ \hline
\textsc{DCVED} & med  & 49.3 / 28.3 / 14.2 & 44.3 & 73.9  \\


\textsc{GPT-C2F} & med &  49.0 / 28.3 / 14.6 & 45.3 & 76.4  \\
\textsc{GPT-TabGen} & med & 49.6 / 28.2 / 14.2 & 44.7 & 74.6  \\
\textsc{DevTC$_{mask}$} & med & \textbf{50.8} / \textbf{29.2} / \textbf{15.2} & \textbf{45.6} & \textbf{77.0}  \\ \hline
\\
& \multicolumn{2}{c}{\textsc{Logic2Text}} & & \\ \hline
Model & Size & BLEU 1/2/3 ($\uparrow$) & SP-A ($\uparrow$) & NLI-A ($\uparrow$)  \\ \hline 

\textsc{DCVED} & med &  46.4 / 31.2 / 20.1 & \textbf{43.7} & 71.9 \\

\textsc{GPT-C2F*} & med & 46.6 / 31.1 / 20.5 &	40.8 & 73.4  \\ 
\textsc{GPT-TabGen*} & med & 46.1 / 32.4 / 21.0 & 41.0 & 70.3  \\ 
\textsc{DevTC$_{mask}$} & med & \textbf{47.8} / \textbf{32.6} / \textbf{22.2} & 41.9 & \textbf{74.4} \\ \hline

\end{tabular}
}

\caption{Quality results on the test split of \textsc{LogicNLG} and \textsc{Logic2Text}. Baseline models trained by us are marked with a \textsc{*}, all \textsc{DevTC} and starred results are the average over 5 different seeds, \textbf{bold} marks statistically significant advantage. \textsc{DevTC} is marked with \textit{mask} to indicate the use a mask token  as the type.}
\label{tab:results_logicnlg}
\end{table} 

\begin{figure}
\includegraphics[width=\columnwidth]{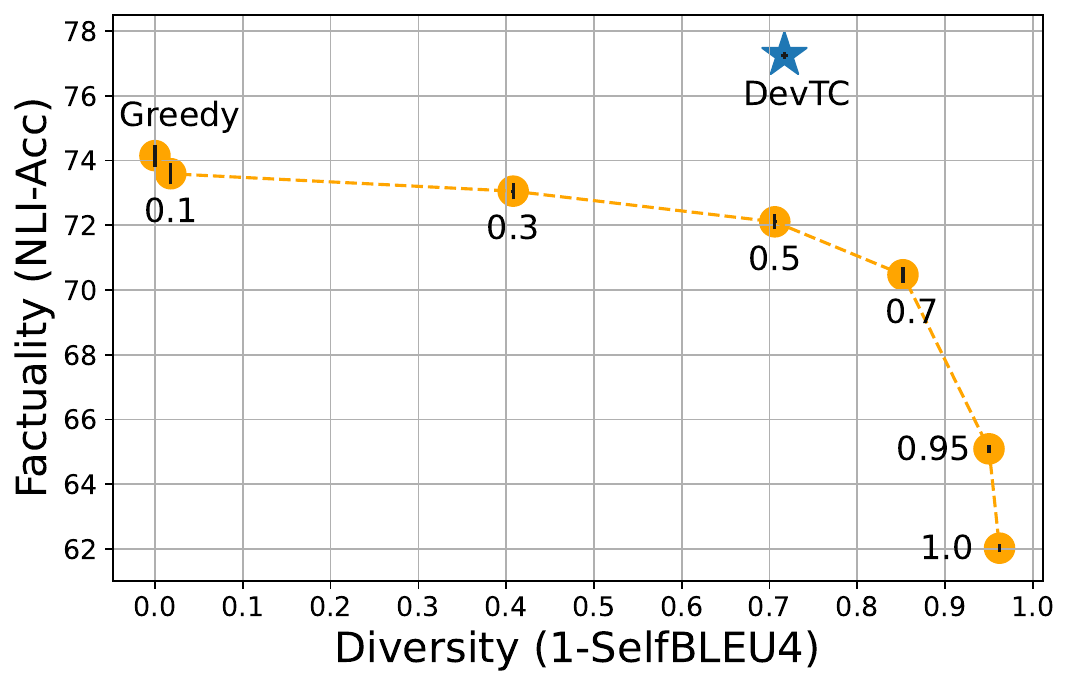}
\caption{Factuality-diversity trade-off for LogicNLG: each dot in the \textcolor{orange}{orange} line represents  an average over 5 seeds (error bars are SEMs) of the baseline model (\textsc{GPT-TabGen}) with nucleus sampling parameters varied between $0$ (greedy decoding) and $1$. 
The \textcolor{blue}{blue} star is our method (using greedy decoding) that surpasses the the baselines pareto frontier.}
\label{fig:tradeoff}
\end{figure}



\subsection{Quality Performance}
\label{Automatic evaluations}

Conventional T2T setup evaluation was done on the \textsc{LogicNLG} and \textsc{Logic2Text} test-sets. As in \cite{chen2021confounded}, when evaluating on \textsc{Logic2Text} we follow the Logical NLG task formulation and do not use the logical-form annotations. 
Since LT annotations are unavailable in this scenario, our type-controlled models are conditioned on a mask token as control. As shown in Table~\ref{tab:results_logicnlg}, for both datasets, across metrics and model sizes \textsc{DevTC} outperforms all baseline methods. 

\subsection{Factuality-Diversity Trade-off}
\label{tradeoff}

To compare \textsc{DevTC} and \textsc{GPT-TabGen} (the strongest baseline) across the Factuality-Diversity plane, we generated a set of $5$ statements per table for each method. 
Since, as opposed to \textsc{DevTC}, which natively enables the production of a diverse set of statements via LT-control, the baseline cannot produce a diverse set with greedy decoding, we utilized stochastic decoding, the most common practice to obtain a set of different outputs from a single model. 
Following \newcite{ippolito2019comparison} we varied the $top_p$ decoding parameter of the baseline to explore the factuality-diversity trade-off for the baseline. 
In contrast, \textsc{DevTC} allows us to achieve diversity without using stochastic decoding (which is known to reduce quality), by conditioning on different LTs.
Fig.~\ref{fig:tradeoff} shows results of \textsc{DevTC} obtained by conditioning on $5$ LTs sampled uniformly from the $7$ LTs, compared to the baseline paired with stochastic decoding as described above.
To evaluate, we measured the diversity within each set, along with the average \textsc{NLI-A}. Figure~\ref{fig:tradeoff} shows that \textsc{DevTC} is better positioned on the factuality-diversity plane, surpassing the baselines Pareto front.
We attribute this gain in generation factuality to the use of more accurate supervision through the LTs, offloading the task of LT prediction from the model, and bypassing the quality degradation incurred by stochastic decoding. 
We found these results to be consistent across other diversity measures such as Ent-2/4 and Dist-2/4, decoding methods, and datasets (see Appendix \ref{A:more tradeoff} for more results). 

\subsection{Human Evaluation}
\subsubsection{Factuality and Diversity}
We complement the automatic evaluation results with human evaluation, to verify the success of our approach in gaining a given diversity with higher factually. 
We therefore choose the $top_p$ decoding parameter of \textsc{GPT-TabGen} to be the one that produces the most similar output to \textsc{DevTC} in terms of diversity (i.e. $0.5$).
We sampled $100$ tables from the set used in Section~\ref{tradeoff} and distribute them independently to 3 human experts.
Each table was presented along with two $5$-statement sets -- one generated by \textsc{DevTC}, and the other by \textsc{GPT-TabGen}. 
The experts were asked which of the two sets is more factual, i.e., properly describes the data in the table (ties are also allowed), and which is more diverse -- on Likert scale, from $-2$ (set-1 is much better) to $+2$ (set-2 is much better). 
Overall, the human evaluation findings are inline with the results appearing in Figure~\ref{fig:tradeoff}.
In $55\%$ of the samples presented to the annotators, \textsc{DevTC} was reported to be more factual compared to $21\%$ for \textsc{GPT-TabGen}. The rest $24\%$  were reported as a tie. 
\textsc{DevTC} advantage is statistically significant ($P_{value}$<$0.05$) using two-sided t-test. 
For diversity, there was a slight advantage to \textsc{DevTC} implying no significant difference inline with Fig~\ref{fig:tradeoff}.
The annotators inter rater Cohen's Kappa is $0.553$ indicating moderate agreement.

\subsubsection{LT-Control}
To verify our models proficiency in LT-control we asked the experts to classify the LTs of the above generated statements. 
The LT-consistency (i.e., the ratio of examples where control LT resulted in a generated statement classified to the same LT) on average over the 7 types is $79.8$\% (for comparison, a baseline model produced a ratio of $17$\%).
The lowest consistency is for \textit{ordinal}, which is characterized with relatively high lexical variance, and for which we had relatively scarce training data.

\section{Conclusions}
\label{conclusions}

We presented \textsc{DevTC}, an innovative model for T2T generation that addresses and implements two prominent features of that task, overlooked by existing models: diversity and control. 
\textsc{DevTC} facilitates the generation of a statement of a desired LT, and the option to generate a diverse set of high quality statements, features that are unlocked by adding statement LT-control to the input.
Results show the merit of our approach compared to existing baselines in generation quality as measured by common benchmarks, diversity-factuality trade-off in automatic and human evaluations. 

\label{limitations}
\section{Limitations}

The main limitations of our work are automatic factuality evaluation and factual generation. In terms of automatic factuality evaluation, current SOTA table fact-checking metrics such as NLI-Acc and SP-Acc still present medium human agreement (See Figure~\ref{fig: examples}). In terms of 
factual generation as determined by human evaluation, 
as all End-to-end T2T methods, \textsc{GPT-TabGen}, the method we use to show the ability of \textsc{DevTC} to improve diversity without sacrificing accuracy, suffers from weak human approval in terms of factuality. As we show in the main text, \textsc{DevTC} is able to improve the factuality over the baseline but still presents human approval factuality that is too low for business applications.


\bibliography{references}
\bibliographystyle{acl_natbib}

\appendix
\label{Appendix}
\section{Appendix}

\subsection{Dataset Statistics}
\label{A: Dataset Statistics}

See Table~\ref{tab:datasets}.

\begin{table}[h!]
\centering
\resizebox{\columnwidth}{!}{%
\begin{tabular}{lccc}
Dataset  & Parent tables & Statements & Train / Dev / Test \\ \hline 
\textsc{LogicNLG}  & 7,392 & 37,015 & 28,450 / 4,260 / 4,305 \\
\textsc{Logic2Text}  & 5,554 & 10,753 & 8,566 / 1,095 / 1,092 \\ 
\end{tabular}
}\caption{Datasets statistics.}
\label{tab:datasets}
\end{table} 

\subsection{Implementation Details}
\label{A: Implementation details}
All models are trained with batch size of 32 on 1 NVIDIA A100 GPUs for 12 epochs. We use Adam optimizer \cite{kingma2014adam} and an autoregressive cross entropy loss to optimize the models. During test time, we use a greedy search to generate text and calculate the BLEU-1,2,3 scores with the 5 references from all 5 sub-tables as suggested by \cite{chen2020logical}. We base our implementation on Huggingface’s Transformers \cite{wolf2019huggingface}  version 4.16.2 in the \cite{paszke2019pytorch} flavour and use the pre-trained version of GPT-2 \cite{radford2019language} small/medium with subword unit vocabulary of 30K. All models selection is based on the BLEU-3 score on dev set. All our models and models marked with a \textsc{*} were found to have the best performance with learning rate set to $1e$-$5$.
Regarding \textsc{DCVED}, we note that, we report the original variant of \textsc{DCVED} without an additional generate-and-select scheme, since multiple generation and re-ranking is complementary and could potentially be applied to all compared methods.


\begin{figure}
\includegraphics[width=\columnwidth]{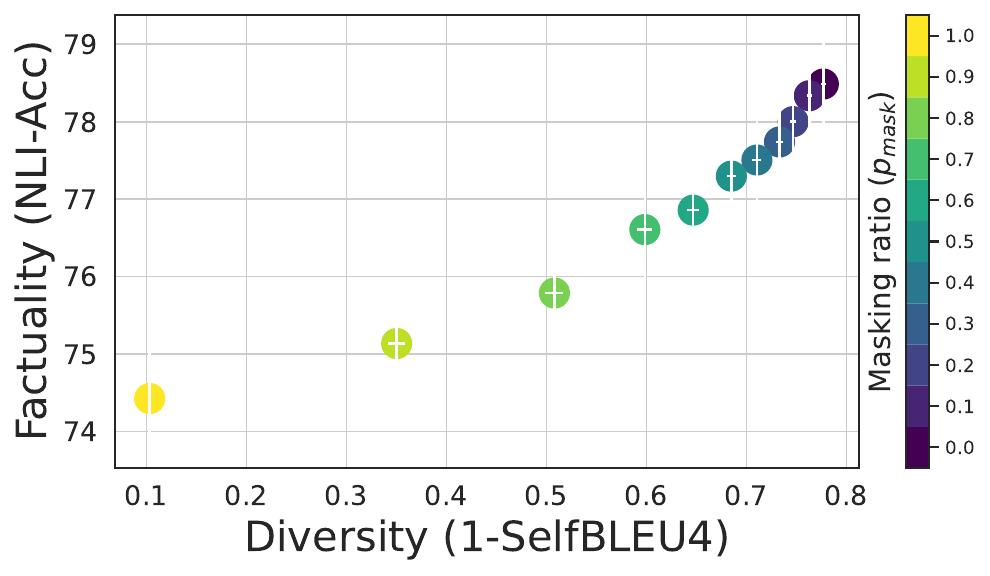}
\caption{Factuality-diversity trade-off over the LogicNLG dataset for different $p_{mask}$, averaged over 5 seed (error-bars are SEMs).} 
\label{fig:ablation}
\end{figure}

\subsection{Masking Ratio Effect}
\label{A :masked_ratio}
To analyze how the different LT masking ratios used in training impact model performance, we trained $11$ LT-controlled models with $p_{mask}$ varying from $0.0$ (no masking) to $1.0$ (always masked). In Figure~\ref{fig:ablation} we compare these models using the same evaluation protocol as in Section~\ref{tradeoff}. 
As expected, both factuality and diversity obtained by \textsc{DevTC} gain significantly from strengthening the control. 
That is, as expected, a lower masking ratio means a more stable training process with better LT correspondence, which in turn results in higher diversity and better factuality on the test set.

Table~\ref{tab:automatic sem} is complementary to the automatic evaluation and includes the standard error of the mean for our models.

\begin{table}[ht!]
\centering
\resizebox{\columnwidth}{!}{ 

\begin{tabular}{lcccc}
& \multicolumn{2}{c}{LogicNLG} & & \\ \hline
Model & Size & BLEU 1/2/3 ($\uparrow$) & SP ($\uparrow$) & NLI ($\uparrow$)  \\ \hline

\textsc{DevTC} & sm & 50.0$\pm$0.2 / 28.6$\pm$0.2 / 14.4$\pm$0.2 & 43.0$\pm$0.3 & 73.4$\pm$0.5 \\ \hline
\textsc{DevTC} (oracle) & sm &  51.3$\pm$0.1 / 30.3$\pm$0.1 / 15.6$\pm$0.1 & 40.5$\pm$0.5 & 75.4$\pm$0.2 \\
& & & & \\

\textsc{DevTC} & med & 50.8$\pm$0.2 / 29.2$\pm$0.2 / 15.2$\pm$0.2 & 45.6$\pm$0.5 & 77.0$\pm$0.6  \\ \hline
\textsc{DevTC} (oracle) & med &  52.3$\pm$0.2 /	31.1$\pm$0.2 / 16.7$\pm$0.2 & 42.7$\pm$0.5 & 78.2$\pm$0.2  \\
\\
& \multicolumn{2}{c}{\textsc{Logic2Text}} & & \\ \hline
Model & Size & BLEU 1/2/3 ($\uparrow$) & SP ($\uparrow$) & NLI ($\uparrow$)  \\ \hline 

\textsc{DevTC} & med & 47.8$\pm$0.2 / 32.6$\pm$0.1 /22.2$\pm$0.1 & 41.9$\pm$0.2 & 74.4$\pm$0.7 \\ \hline
\textsc{DevTC} (oracle) & med &  48.4$\pm$0.2 / 33.6$\pm$0.2 / 23.2$\pm$0.1 & 42.6$\pm$0.7  & 76.1$\pm$0.5 \\

\end{tabular}}

\caption{Quality results on the test split of \textsc{LogicNLG} and \textsc{Logic2Text}, all \textsc{DevTC} results are the average over 5 different seeds, the $\pm$s represents the standard error of the mean.}
\label{tab:automatic sem}

\end{table} 


\begin{figure*}[th!]
\includegraphics[width=\textwidth]{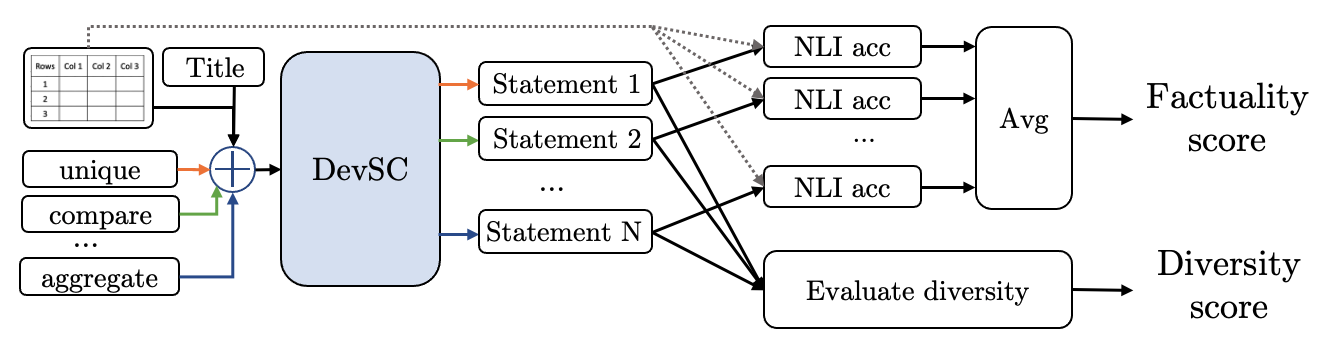}
\caption{An illustration of the quality-diversity trade-off evaluation. NLI-Acc is a fact checking model proposed by \newcite{chen2020logical} that labels the statement as true or false given the table.}
\label{fig:tradeoff evaluation}
\end{figure*}

\subsection{Factuality-Diversity Trade-off: Other diversity measures}
\label{A:more tradeoff}

\begin{figure*}[ht!]
\center

\makebox[\textwidth][c]{
  \includegraphics[width=0.5\textwidth]{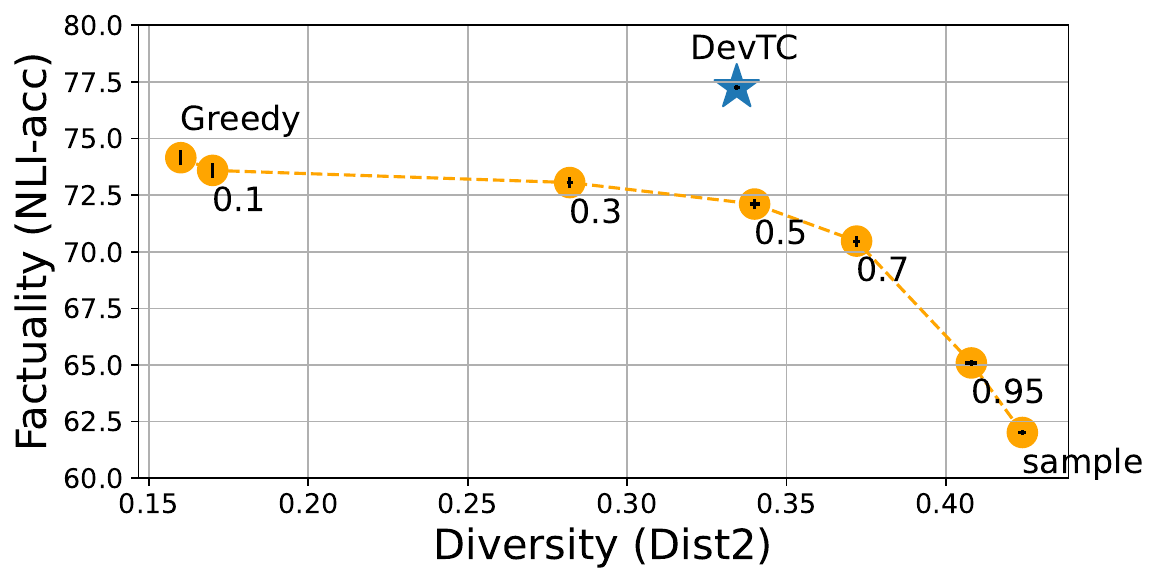}
  \includegraphics[width=0.5\textwidth]{div_tradeoff_NLI_SB4.pdf}
}

\caption{Factuality-Diversity trade-off for Dist-2 and Self-BLEU4: each dot in the \textcolor{orange}{orange} line represents  an average over 5 seeds (error bars are SEMs) of the baseline model (\textsc{GPT-TabGen*}) with a different nucleus sampling decoding parameters (shown in the figure). The \textcolor{blue}{blue} star is our method that surpasses the trade-off line created by the baseline and the decoding strategy.}
\label{fig:more_tradeoff}

\end{figure*}

Figure~\ref{fig:more_tradeoff} displays the factuality-diversity trade-off discussed in Section~\ref{tradeoff} for the other two diversity metrics, SelfBLEU4 and Dist2.

\begin{figure*}[ht!]
\includegraphics[width=\textwidth]{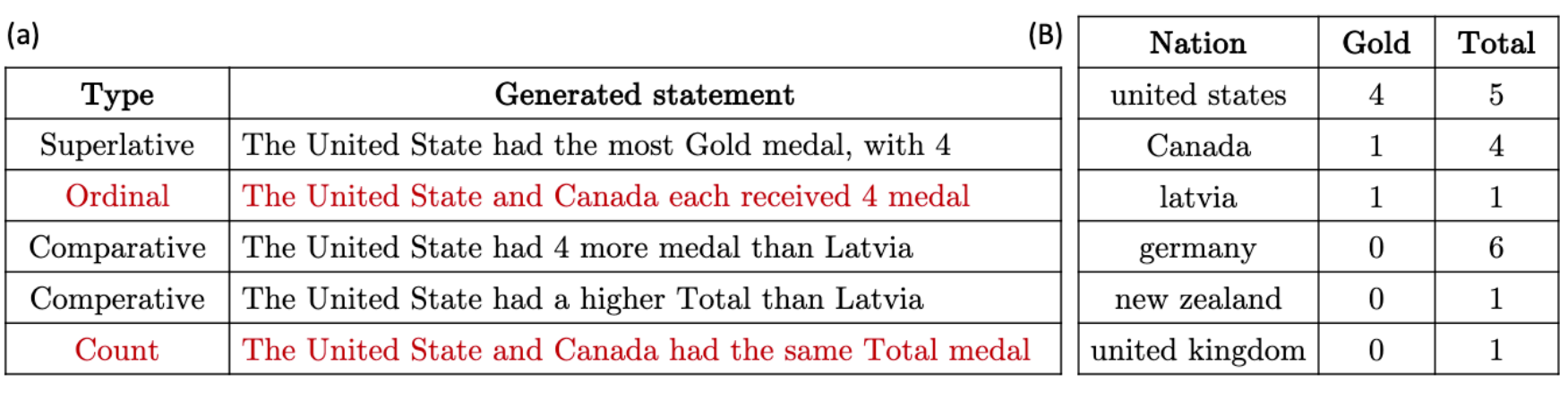}
\caption{5 statements generated using \textsc{DevTC} along with the table that was used for their generation, sentences marked in red display false type correspondence.}
\label{fig: examples}
\end{figure*}


\end{document}